%% file: main.tex
\documentclass[letterpaper, 10 pt, conference]{ieeeconf}  %

\IEEEoverridecommandlockouts                              %

\input{_macros}

\title{Gondola \includegraphics[height=25pt]{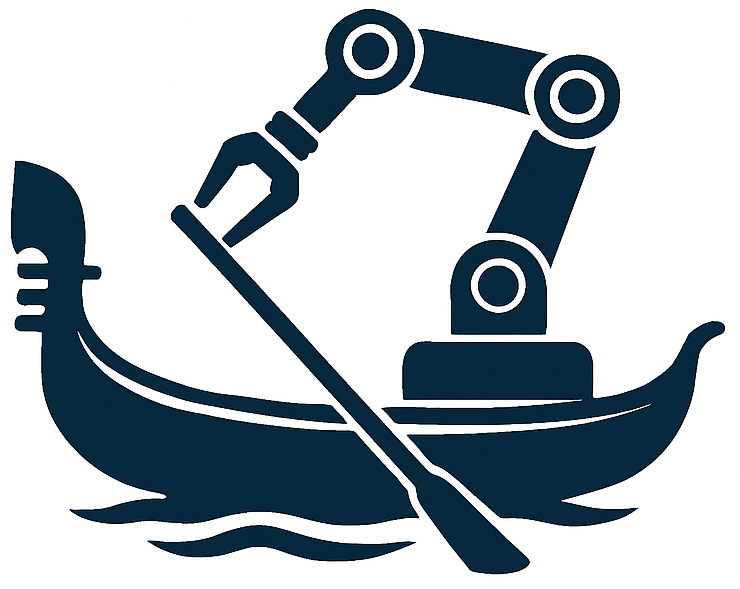}: Grounded Vision Language Planning for Robotic Manipulation}

\author{Shizhe Chen$^1$, Ricardo Garcia$^1$, Paul Pacaud$^1$, and Cordelia Schmid$^1$\\
\thanks{$^1$Inria, \'Ecole normale sup\'erieure, CNRS, PSL Research University, 75005, Paris, France. {\tt\small \{firstname.lastname\}@inria.fr}}
}

\begin{document}

\maketitle
\thispagestyle{empty}
\pagestyle{empty}

\input{00_abstract}

\input{01_intro}

\input{02_related}

\input{03_method}

\input{04_experiments}

\input{10_conclusion}

{
\small
\textbf{Acknowledgements.}
This work was granted access to HPC resources of IDRIS under the allocation AD011014846 made by GENCI. 
It was funded in part by the French government under management of Agence Nationale de la Recherche as part of the “France 2030" program, reference ANR-23-IACL-0008 (PR[AI]RIE-PSAI projet), the ANR project 3D-GEM (ANR-25-CE23-7777-01), the ANR project VideoPredict (ANR-21-FAI1-0002-01) and the Paris Île-de-France Région in the frame of the DIM AI4IDF. 
Cordelia Schmid would like to acknowledge the support by the Körber European Science Prize.
}

\bibliographystyle{IEEEtran}
\bibliography{11_references}

\end{document}

%% file: _macros.tex
\usepackage{graphicx}	
\usepackage{amsmath}	
\usepackage{amssymb}	
\usepackage{booktabs}
\usepackage{times}
\usepackage{microtype}
\usepackage{epsfig}
\usepackage{caption}
\usepackage{float}
\usepackage{placeins}
\usepackage{color, colortbl}
\usepackage{stfloats}
\let\labelindent\relax
\usepackage{enumitem}
\usepackage{tabularx}
\usepackage{xstring}
\usepackage{multirow}
\usepackage{xspace}
\usepackage{url}
\usepackage{subcaption}
\usepackage{xcolor}
\usepackage{tcolorbox}
\usepackage[hang,flushmargin]{footmisc}
\usepackage{wrapfig} 
\usepackage{xurl}
\usepackage[colorlinks=true,breaklinks=true]{hyperref}

\newcommand{\R}[1]{{%
    \textbf{%
        \ifstrequal{#1}{1}{\textcolor{red}{R#1}}{%
        \ifstrequal{#1}{2}{\textcolor{blue}{R#1}}{%
        \ifstrequal{#1}{3}{\textcolor{magenta}{R#1}}{%
        \ifstrequal{#1}{4}{\textcolor{teal}{R#1}}{%
                           \textcolor{cyan}{R#1}%
        }}}}%
    }%
}}

%% file: 00_abstract.tex
\begin{abstract}
Vision-language-action (VLA) models have shown promising progress in robotic manipulation. However, directly mapping visual observations and language instructions to low-level actions often results in limited interpretability and weak robustness in complex, long-horizon tasks.
To address these challenges, we employ a modular manipulation framework that separates high-level planning from low-level control. 
At its core is \emph{Gondola}, a grounded vision-language planning model that generates structured plans with explicit pixel-level object grounding before action execution. Given multi-view observations and planning history, Gondola predicts the next-step plan as interleaved textual instructions and multi-view segmentation masks corresponding to target objects and goal locations. 
To train Gondola, we construct synthetic datasets that provide explicit supervision for short-horizon grounded planning, multi-view referring expression, and long-horizon compositional reasoning. 
By coupling grounded plan generation with a 3D-based execution policy, our framework achieves state-of-the-art performance on the challenging GemBench benchmark.
The system further demonstrates promising transfer to real robots.
Ablation studies confirm that pixel-level grounding and the proposed planning-oriented supervision are critical for effective high-level reasoning.
Project webpage: \url{https://cshizhe.github.io/projects/robot_gondola.html}

\end{abstract}

%% file: 01_intro.tex
\section{Introduction}
\label{sec:intro}

Training robots to perform physical manipulation tasks following human instructions has been a long-term goal in robotics, enabling intuitive human-robot interaction in unstructured, dynamic environments such as homes and factories. 
Recent end-to-end learning-based approaches~\cite{chi2023diffusionpolicy,fu2024mobilealoha}, particularly Vision-Language-Action models (VLAs)~\cite{brohan2023rt2,team2024octo,kim2024openvla,qu2025eo1,bjorck2025gr00t,black2024pi_0} trained on real-world robot data, have achieved impressive progress in robotic manipulation. 

Despite these advances, directly mapping high-dimensional visual observations and language instructions to low-level continuous actions remains fundamentally challenging. Such end-to-end formulations entangle perception, semantic reasoning, and control into a single model, making them data-hungry, difficult to interpret, and sensitive to distribution shifts. Due to the scarcity and limited diversity of available robot datasets~\cite{vuong2023openx,khazatsky2024droid,contributors2025agibotworld}, these models often struggle to generalize beyond their training conditions, particularly when encountering novel objects, unfamiliar environments, or unseen long-horizon task compositions~\cite{mees2022calvin,pumacay2024colosseum,garcia2024gembench}.

\begin{figure}[t]
    \centering
     \includegraphics[width=\linewidth]{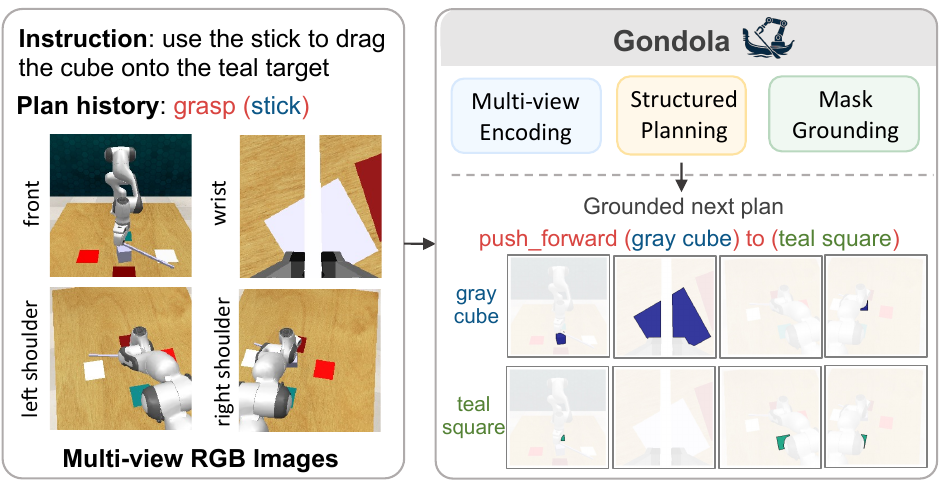}
     \caption{Gondola simultaneously produces an interpretable, mask-grounded structured plan that localizes task-relevant objects and placements across views.}
    \label{fig:teaser}
    \vspace{-1.em}
\end{figure}

To improve generalization, modular frameworks~\cite{liang2023codepolicy,huang2023voxposer,garcia2024gembench,bjorck2025gr00t} have gained increasing attention by separating high-level task planning from low-level action execution. 
Decoupling planning from embodiment allows the planner to leverage large-scale internet data and pretrained foundation models, potentially improving compositional reasoning and task-level generalization.
Earlier works employ Large Language Models (LLMs)~\cite{openai2023gpt4v} for high-level task planning~\cite{garcia2024gembench,huang2022languageplanner,liang2023codepolicy}. However, these planners often lack direct grounding in visual observations, limiting their ability to produce physically actionable plans. 
Several approaches attempt to bridge this gap by introducing affordance scoring of candidate plans~\cite{brohan2023saycan} or connecting language planning with additional perception APIs~\cite{liang2023codepolicy}. 
While these methods improve alignment between language and execution, they rely on predefined perception and skill sets, which may omit critical visual information and propagate errors.

More recently, several works~\cite{li2024llara,yuan2024robopoint,bjorck2025gr00t} have explored adapting Vision-Language Models (VLMs) to generate visually grounded plans, using intermediate representations such as points~\cite{yuan2024robopoint}, bounding boxes~\cite{li2024llara}, or latent hidden states from VLMs~\cite{bjorck2025gr00t}.
While promising, points and bounding boxes are often too coarse for precise robotic manipulation in 3D environments. 
Latent representations, on the other hand, are difficult to interpret and may lack the conciseness needed for efficient execution.
A line of efforts has investigated pixel-level grounding with VLMs~\cite{lai2024lisa,rasheed2024glamm,yuan2025sa2va}. However, these works primarily focus on general image understanding, rather than integrating grounded representations into robotic planning. 
To the best of our knowledge, such grounded VLMs have not yet been explored for high-level task planning.
Moreover, existing grounding approaches rely on single-view observations, which restrict spatial context and exacerbate ambiguity under occlusions.

To address these limitations, we propose \textbf{Gondola}, a \textbf{g}r\textbf{o}u\textbf{n}de\textbf{d} visi\textbf{o}n–\textbf{l}anguage pl\textbf{a}nning model that generates executable high-level plans together with explicit pixel-level object grounding. As illustrated in Figure~\ref{fig:teaser}, Gondola transforms language instructions, multi-view observations, and prior executed actions into structured next-step plans composed of interleaved action descriptions and object or location references, accompanied by segmentation masks for the referred entities.
The unified text and mask generation is enabled by incorporating a special segmentation token~\cite{lai2024lisa,rasheed2024glamm,yuan2025sa2va} into the VLM backbone, allowing object-specific mask prediction across views.
This grounded plan representation forms an interpretable and spatially precise interface between semantic reasoning and physical execution.
For training our model, we construct three types of supervised data in simulated RLBench environments~\cite{james2020rlbench}: short-horizon grounded planning data for step-wise plan supervision, multi-view referring expression data to strengthen cross-view object grounding, and pseudo long-horizon grounded planning data to improve compositional reasoning. To generate the long-horizon data, we synthesize extended tasks by concatenating shorter task sequences and leveraging an LLM to produce coherent task instructions.
We evaluate Gondola on both offline grounded plan generation and online task execution using the GemBench benchmark~\cite{garcia2024gembench}.
For online execution, Gondola is integrated with a 3D-based motion policy~\cite{garcia2024gembench} that executes the generated grounded plans in a closed-loop manner.
Extensive experiments demonstrate that multi-view pixel-level grounding and planning-oriented supervision are all critical for robust high-level reasoning. 
Gondola achieves state-of-the-art performance on GemBench and shows promising transfer to real-world robotic manipulation.

To summarize, our contributions are three-fold:
\parskip=0.1em
\begin{itemize}[itemsep=0.1em,parsep=0em,topsep=0em,partopsep=0em]
    \item We decouple high-level reasoning from low-level control for robotic manipulation, and propose \textbf{Gondola} to generate structured, mask-grounded plans as an explicit interface between perception and action. %
    \item We construct three complementary training datasets, including short-horizon grounded planning, multi-view referring expressions, and pseudo long-horizon tasks, to explicitly supervise grounded plan generation and enhance long-horizon compositionality.
    \item Gondola outperforms prior VLMs~\cite{yuan2024robopoint,lai2024lisa} on grounded task planning with unified multi-view pixel-level segmentation and planning-oriented supervision. When integrated with a 3D-based execution policy, it achieves the state-of-the-art  performance on GemBench. %
\end{itemize}
The code, models and datasets will be publicly released.

%% file: 02_related.tex
\section{Related Work}
\label{sec:related}

\noindent\textbf{Vision-language robotic manipulation.}
Learning robotic manipulation conditioned on vision and language has garnered significant interest~\cite{shao2021concept2robot,lynch2023interactivelang,stepputtis2020language}. 
Due to the high-dimensional action space, directly applying reinforcement learning (RL) for training presents challenges~\cite{kalashnikov2018scalablerl}. Therefore, most approaches employ imitation learning~(IL)~\cite{jang2022bcz,brohan2022rt1,guhur2023hiveformer,shridhar2023peract,goyal2023rvt,chen2023polarnet,chen2024sugar,gervet2023act3d,ke20243ddifusseractor} using scripted trajectories~\cite{james2020rlbench} or tele-operation data~\cite{vuong2023openx}. 
Visual representation plays a crucial role in policy learning. Existing works~\cite{jang2022bcz,brohan2022rt1,chi2023diffusionpolicy,guhur2023hiveformer,goyal2023rvt,goyal2024rvt2, tziafas2024towards} rely on 2D images for action prediction, although recent work explores 3D visual representations~\cite{james2022c2farm,shridhar2023peract,chen2023polarnet,gervet2023act3d,ke20243ddifusseractor,chen2024sugar,garcia2024gembench, chisari2024learning}.
Hiveformer~\cite{guhur2023hiveformer} and RVT~\cite{goyal2023rvt} utilize 2D images to predict a heatmap in 2D space, which is then combined with 3D point clouds to demermine the final 3D position.
C2F-ARM~\cite{james2022c2farm} and PerAct~\cite{shridhar2023peract} directly use 3D voxel representation as input, being less efficient due to encoding empty voxels.
PolarNet~\cite{chen2023polarnet} and 3D-LOTUS~\cite{garcia2024gembench} improve efficiency by encoding only visible point clouds to predict actions, while SUGAR~\cite{chen2024sugar} further enhances point cloud representation through 3D pre-training.
Given the superiority of current pre-trained 2D representations~\cite{radford2021clip}, works like Act3D~\cite{gervet2023act3d} and 3D Diffuser Actor~\cite{ke20243ddifusseractor} lift pre-trained 2D features into 3D space, and then train 3D models to leverage the strengths of both.
In this work, we leverage the strong capabilities of 2D VLMs for high-level task planning and integrate it with 3D-based motion planning policies for task execution.

\noindent\textbf{Foundation models for robotics.}
Learning-based robotic policies struggle to generalize to novel scenarios~\cite{yu2023scaling}.
nspired
by generalization abilities of foundation models~\cite{radford2021clip,ravi2024sam2,openai2023gpt4v}, recent research leverages these models for perception, reasoning and planning in robotics.
A line of work, commonoly referred to as VLAs~\cite{kim2024openvla,qu2025eo1,bjorck2025gr00t,black2024pi_0}, directly fine-tunes VLMs to predict low-level action sequences, while another line of work uses foundation models in a modular manner. 
Some methods~\cite{huang2022languageplanner,garcia2024gembench} use LLMs to decompose high-level tasks into sub-steps. 
To better ground plans in vision, SayCan~\cite{brohan2023saycan} combines LLMs with value functions of pretrained skills given visual contexts. ViLa~\cite{hu2023vila} replaces LLMs with a VLM GPT-4V~\cite{openai2023gpt4v}. CaP~\cite{liang2023codepolicy} directs LLMs to generate code that invokes tools for visual perception and control, and VoxPoser~\cite{huang2023voxposer} uses LLMs and VLMs to create 3D voxel maps. 
These approaches rely on general-purpose pretrained models for task planning, but tend to be unstable in robotic settings and require heavy prompt engineering.
To address this, a few methods fine-tune VLMs on embodied datasets for better grounding, with points~\cite{yuan2024robopoint} or bounding boxes~\cite{li2024llara}.
In our work, we extend the VLM framework with multi-view inputs and finer grounding masks, and introduce synthetic robot data for long-horizon reasoning.

\noindent\textbf{Vision and language models for grounding.}
Early VLMs~\cite{liu2024llava,liu2024llava1.5} are constrained to generating text outputs from multimodal inputs, such as image captioning and visual question answering. 
To enable VLMs to produce grounded outputs that align generated texts with specific image regions, existing methods can be broadly categorized into three types. 
The first category outputs box coordinates~\cite{peng2023kosmos2,chen2023shikra,you2023ferret} or polygons of segmentation masks~\cite{liu2023polyformer} as text. However, generating precise numerical outputs 
is difficult and prone to hallucination.
The second category uses a proposal-based approach, where a separate module first generates candidate regions, and the VLM selects the one for each generated text~\cite{ma2025groma,zhang2024groundhog}. 
While more structured, this method is sensitive to proposal quality, suffers from error accumulation, and introduces more computation overhead.
The third category decouples language and grounding by feeding the output of a VLM into a grounding model to produce boxes or masks~\cite{lai2024lisa,rasheed2024glamm,yuan2025sa2va} via a special segmentation token. 
Our Gondola model extends pretrained grounding VLMs~\cite{lai2024lisa,yuan2025sa2va} to multi-view image grounding and planning for robotic manipulation. %

%% file: 03_method.tex
\section{Gondola: Grounded Vision-Language Planning Model}
\label{sec:method}

\begin{figure*}[tp]
    \centering
    \includegraphics[width=\linewidth]{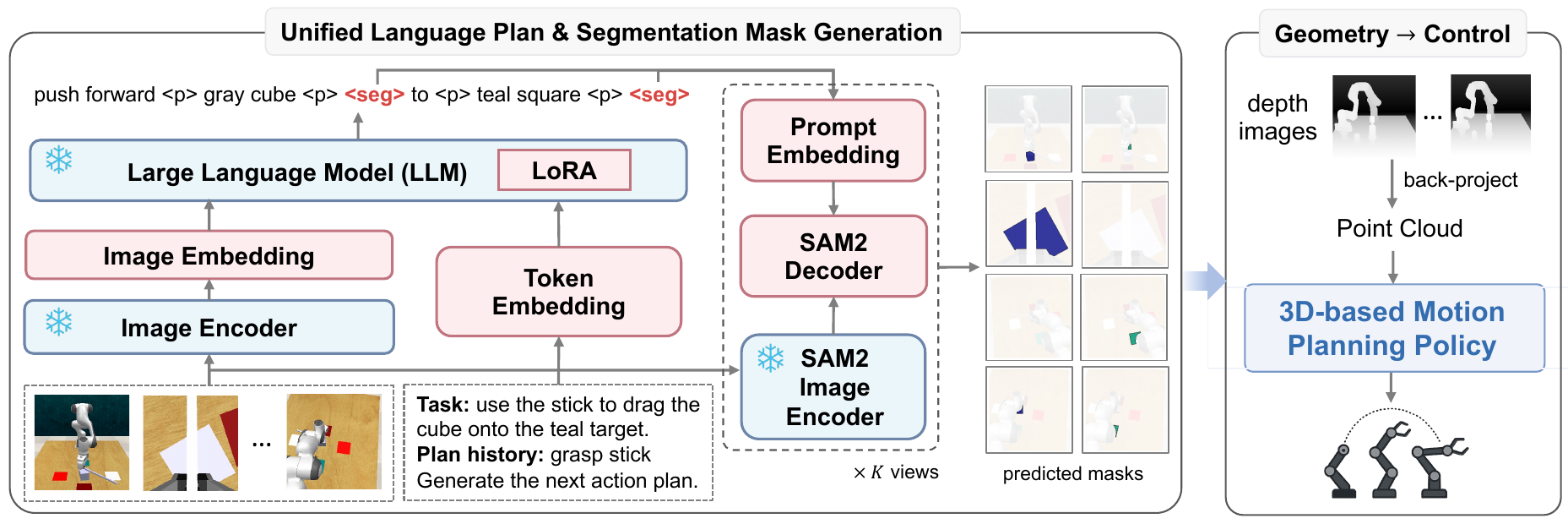}
    \caption{Left: Gondola model architecture, consisting of a shared visual encoder for multi-view images, an LLM to generate action and object names along with segmentation tokens, and SAM2 to decode masks. Right: Integrating Gondola with a 3D-based motion planning policy for closed-loop task execution.}
    \label{fig:method}
    \vspace{-1.em}
\end{figure*}

We formulate high-level task planning for robotic manipulation as a vision-language grounding problem, where the goal is to generate the next executable, visually-grounded action plan that accomplishes a natural language instruction in the observed environment. 
Formally, given a language instruction $L$ and multi-view visual observations $I = \{I_1, \cdots, I_K\}$ from $K$ cameras, the Gondola model produces a grounded vision-language plan $P = (a, o, M_o, l, M_l)$, where $a$ represents the action name, $o$ specifies the manipulated object description paired with corresponding segmentation masks $M_o = \{m_o^1, \cdots, m_o^K\}$  across all views, and $l$ denotes the target location description with associated location masks $M_l$.
Noting that either $o$ or $l$ may be empty if the particular action does not require an object or target location for execution.

\subsection{Model Architecture}
\label{sec:method_gondola}

As illustrated in Figure~\ref{fig:method}, Gondola consists of three main components: an image encoder for tokenizing each view image, an LLM to process multimodal inputs and outputs, and a segmentation model for multi-view object grounding.

\noindent\textbf{Image encoder.} 
We use a pretrained vision transformer (ViT) InternVL-300M~\cite{chen2024internvl} with image resolution of 448x448 to generate image patch embeddings, followed by a 2-layer multi-layer perceptron (MLP).
The ViT is frozen, while the MLP is trained to adapt the visual features to the language space.
The same image encoder is applied across all views, with view separation handled by a special token \verb|\n|. All image tokens are concatenated to form a single sequence.

\noindent\textbf{LLM.} 
We adopt InternVL-4B~\cite{chen2024internvl} as our language model, keeping its base parameters frozen while adding LoRA~\cite{hu2021lora} layers for fine-tuning.
Inspired by grounding VLMs~\cite{lai2024lisa,yuan2025sa2va}, we incorporate a specialized vocabulary that includes a dedicated \verb|<seg>| token to signal mask generation, along with paired delimiter tokens \verb|<p>| and \verb|</p>| that precisely delineate object and location references requiring spatial grounding.
To maintain contextual awareness across sequential steps in completing manipulation tasks, we further encode previously generated history plans $H$ as compact text tokens to the model.
The following example demonstrates input and output token formatting for the LLM:

\vspace{1mm}
\noindent
\begin{minipage}{\linewidth}
\begin{tcolorbox} 
\small
\textbf{User}: \texttt{<image>}\verb|\n|\texttt{<image>}\verb|\n|\texttt{<image>}\verb|\n|\texttt{<image>}\verb|\n| You are a skilled assistant for robot task planning in tabletop environments. You can perform the following actions: grasp, move\_grasped\_object, rotate\_grasped\_object, push\_down, push\_forward, and release. Task: use the stick to drag the cube onto the teal target. You have completed the following action plans: grasp stick. Please generate the next action plan.\\
\textbf{Gondola}: push forward \texttt{<p>} gray cube \texttt{</p>} \texttt{<seg>} to \texttt{<p>} teal square \texttt{</p>} \texttt{<seg>}.
\end{tcolorbox}
\vspace{0.5mm}
\end{minipage}
Here, \texttt{<image>} represents placeholders for image tokens for each view, which are replaced by the actual visual embeddings.

\noindent\textbf{Segmentation model.}
We employ SAM2~\cite{ravi2024sam2} as the segmentation model.
Given the hidden embedding $h_{\text{seg}}$ from the LLM that predicts the \verb|<seg>| token, we project $h_{\text{seg}}$ with a 2-layer MLP to generate a prompt embedding.
SAM2 uses this prompt to segment the corresponding object mask for each view image separately and thus generates $K$ binary masks for each referred object or location.

\subsection{Training Data}
\label{sec:method_data}

\begin{figure*}
    \centering
    \includegraphics[width=1\linewidth]{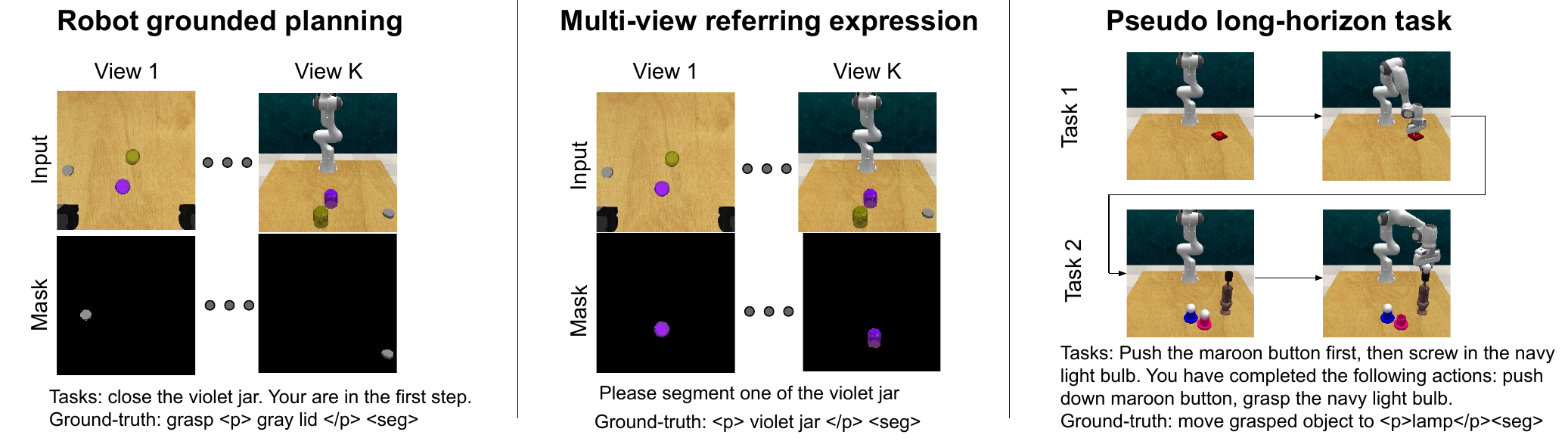}
    \vspace{-1.5em}
    \caption{Constructed datasets: (1) robot grounded planning, (2) multi-view referring expressions for improved object grounding, and (3) pseudo long-horizon tasks for enhanced long-horizon reasoning.}
    \label{fig:dataset}
    \vspace{-1.em}
\end{figure*}

We construct three training datasets based on 31 task variations in the GemBench training split~\cite{garcia2024gembench}, including robot grounded planning, multi-view referring expression, and pseudo long-horizon tasks.
While this data construction approach can be extended to any task in RLBench, we restrict dataset construction to the GemBench training split for fair comparison with prior work~\cite{garcia2024gembench}.
Figure~\ref{fig:dataset} illustrates examples from each of the three datasets.

\noindent \textbf{Robot grounded planning.}
In RLBench, each task is structured with semantically labeled objects and fixed procedure trajectories, enabling efficient grounded plan generation.
First, we manually decompose the trajectory in each task into a sequence of plans, each step consisting of an action, object and placement location triplet. This only requires a single annotation effort per task with minimal annotation overhead.
The corresponding segmentation masks for objects and locations are then automatically extracted given the annotated semantic labels.
In this way, we create a ground-truth plan $\{a_t, o_t, M^t_o, l_t, M^t_l\}$ for multi-view images $I_t$ at each keystep
$t$ in an episode per task variation.
We use 100 episodes for each GemBench training task variation, where each episode contains randomized object placements (and optionally new distractor objects), resulting in approximately 15k training tuples for robot grounded planning.

\noindent \textbf{Multi-view referring expression.}
To strengthen multi-view object grounding, we create a multi-view referring expression dataset based on RLBench.
Similar to grounded planning generation, we automatically extract all object instances and their corresponding segmentation masks for each keystep in GemBench training split, yielding 15k multi-view image examples and 58k referring expressions.
For each example, we formulate the referring query as ``Please segment one of the [object name]" with the expected output being the corresponding segmentation masks across all input images.

\noindent \textbf{Pseudo long-horizon tasks.}
To enhance planning for long-horizon tasks, we propose to automatically generate pseudo long-horizon sequences. Specifically, we randomly concatenate pairs of different training task sequences from GemBench training split to create compositional tasks. We then use an LLM to create coherent joint instructions for the combined tasks. 
Despite the abrupt scene transitions between tasks, these pseudo long-horizon tasks still help the model learn to leverage history plans to track task progress, and predict the next step based on long-horizon context.

\subsection{Training Objectives}
\label{sec:method_training}

Gondola is trained to jointly optimize plan generation and multi-view object grounding.
For plan generation, we employ the cross-entropy loss for next token prediction:
\begin{equation}
    \mathcal{L}_{\text{plan}} = -\sum \log p(y_i | y_{<i}, I, L, H),
\end{equation}
where $y_i$ represents tokens in the generated plan including the special tokens.
For multi-view object grounding, we adopt a joint loss of binary mask prediction and dice loss $\mathcal{L}_{\text{grd}} = \mathcal{L}_{\text{bce}} + \mathcal{L}_{\text{dice}}$:
\begin{equation}
\small
    \mathcal{L}_{\text{bce}} = - \sum [M_{\text{gt}}(p) \cdot \log(M_{\text{pd}}(p)) + (1-M_{\text{gt}}(p)) \cdot \log(1-M_{\text{pd}}(p))],
\end{equation}
\begin{equation}
\small
    \mathcal{L}_{\text{dice}} = 1 - \frac{2 \sum_{p} M_{\text{pd}}(p) \cdot M_{\text{gt}}(p)}{\sum_{p} M_{\text{pd}}(p) + \sum_{p} M_{\text{gt}}(p) + \epsilon},
\end{equation}
where $p$ indexes over all pixels in the mask, $M_{\text{pd}}(p)$ is the predicted probability, $M_{\text{gt}}(p)$ is the ground-truth binary label, and $\epsilon$ is a small constant for numerical stability.

\subsection{Closed-Loop Planning-Execution Framework}
We integrate Gondola with a state-of-the-art 3D-based motion policy 3D-LOTUS~\cite{garcia2024gembench} within a closed-loop hierarchical framework. 
In this setting, Gondola serves as a grounded task planner that processes multi-view observations, instruction and previously executed history plans to synthesize the next plan, which includes both the semantic action and pixel-level segmentation masks.
Then we combine aligned depth images with these masks and unify the segmented objects across views into a consolidated 3D point cloud.
Each point is assigned with one of four categories based on the segmentation results and robot proprioceptive information, namely target object, target location, robot, and obstacle. 
The predicted action name and the point cloud are fed into 3DLotus~\cite{garcia2024gembench} to generate a sequence of actions via action chunking. 
After a fixed number of execution steps, the system re-runs the entire pipeline, generating a new plan and action sequences. 
This closed-loop framework enables continuous re-planning and corrections as needed.
While frequent re-planning enhances performance in fine-grained manipulation, it can degrade success rates in long-horizon tasks due to temporal instability over extended sequences (Table~\ref{tab:gembench_online_eval}).

%% file: 04_experiments.tex
\section{Experiments}
\label{sec:expr}

\input{tables/gembench_offline_eval}

\input{tables/gembench_offline_eval_data}

\subsection{Evaluation Datasets and Metrics}

We evaluate Gondola on the GemBench benchmark~\cite{garcia2024gembench} for robotic manipulation in the RLBench~\cite{james2020rlbench} simulator. 
GemBench assesses models' capabilities across four levels: Level 1 (L1) with new locations, Level 2 (L2) with novel rigid objects, Level 3 (L3) with new articulated objects, and Level 4 (L4) with unseen long-horizon tasks.
Only tasks from L1 are used to train the models. The benchmark includes 31 task variations in L1, 28 in L2, 21 in L3 and 12 in L4.
We conduct the following two types of evaluation:

$\bullet$ 
\textbf{Grounded planning evaluation.}
This setup purely assesses models' grounded planning performance given instruction, multi-view images and ground-truth history plans.
We construct a grounded planning evaluation set given the GemBench validation split. It contains 20 episodes per task variation in GemBench. For each keystep in an episode, we provide ground-truth annotations for the next action, object names and segmentation masks across all views.
To evaluate the grounded planning performance, we measure the accuracy of action and object name predictions through exact text matches for each keystep. 
For grounding evaluation, we calculate the intersection over union (IoU) between predicted and ground-truth masks for each view.
The averaged performances of each metric on all keysteps are reported. %

$\bullet$ 
\textbf{Task completion evaluation.}
We adopt the standard camera configuration in GemBench, using $K=4$ cameras positioned at the front, left shoulder, right shoulder and wrist, each with an image resolution of $256 \times 256$.
For evaluation, we use the GemBench test split across all four levels, conducting 20 episodes per task variation for 5 times, resulting in $20 \times 5 \times (31+28+21+12)$ evaluation episodes in total. Each episode is limited to a maximum of 25 steps.
Task performance is measured by success rate (SR), where SR is 1 for a successful episode and 0 for failure.
We report mean SR and standard deviations across the 5 runs.

\subsection{Implementation Details}

We train the Gondola model using 8 NVIDIA H100 GPUs. %
The image encoder and SAM2 encoder are kept frozen during training. We apply LoRA~\cite{hu2021lora} with rank 128 for parameter-efficient fine-tuning of the LLM, and the SAM2 mask decoder is also fine-tuned.
The model is optimized with AdamW, using a learning rate of $2 \times 10^{-5}$ for all trainable parameters. 
The batch size per device is set to 4, resulting in an effective batch size of 32.
It takes 3 hours for training 10k iterations over the three constructed datasets.

\subsection{Grounded Planning Evaluation}

\subsubsection{Ablations}
We perform comprehensive ablations to evaluate the design choices of Gondola, including pixel-level mask grounding, multi-view inputs and history-aware planning.
We further analyze the contribution and scalability of the proposed fine-tuning datasets to understand their effect on grounding quality. 

\noindent\textbf{Boxes vs. Masks.} 
We compare Gondola's mask-based grounding approach with a box-based variant.
The box variant (row 1) in Table~\ref{tab:gembench_offline_eval} directly generates bounding boxes as textual outputs and the predicted boxes are fed into the SAM2 model to produce segmentation masks.
Compared to directly generating masks (row 4) it performs worse across all levels in both action and object name accuracy, and shows significantly lower grounding quality in terms of mask IoU.
We observe that the box-based model frequently suffers from format errors, as generating multiple numeric values for multiple images can be challenging.
These results highlight the advantages of end-to-end mask generation within VLMs, which provides more accurate and reliable grounding for robotic planning.

\noindent\textbf{Multi-view inputs.}
The comparison between the 2nd and 3rd rows in Table~\ref{tab:gembench_offline_eval} showcases the impact of multi-view image inputs for robot task planning.
In the 2nd row,  only the front-view image is provided to the model, whereas in the 3rd row, all four views are used.
Multi-view images help mitigate occlusions and generally improve action, object and grounding prediction across levels.

\noindent\textbf{History plans.}
The last two rows in Table~\ref{tab:gembench_offline_eval} compare the effect of incorporating history plans into task planning.
Including history information boosts the performance on L2 and L3 by enabling more coherent and context-aware planning decisions.
However, on L4, we observe a significant performance drop compared to the model without history. 
An in-depth analysis reveals that this decline is due to a distribution shift in history plans. As a result, the model tends to leverage its prior knowledge for generating purely textual plans rather than grounded plans.
In contrast, the model without history does not suffer from this distribution shift. 
This issue can be addressed by training on our constructed pseudo long-horizon data, as shown in 
Table~\ref{tab:gembench_offline_eval_data}.

\noindent\textbf{Fine-tuning datasets.}
Table~\ref{tab:gembench_offline_eval_data} evaluates the contribution of each fine-tuning dataset.
The multi-view referring expression dataset proves most effective in improving segmentation quality, leading to consistently better grounding performance across all four levels.
The pseudo long-horizon task dataset is particularly beneficial for L4, as it mitigates the history plan shift issue and encourages the model to reason over extended plan histories.

\begin{figure}
  \begin{center}
  \includegraphics[width=0.65\linewidth]{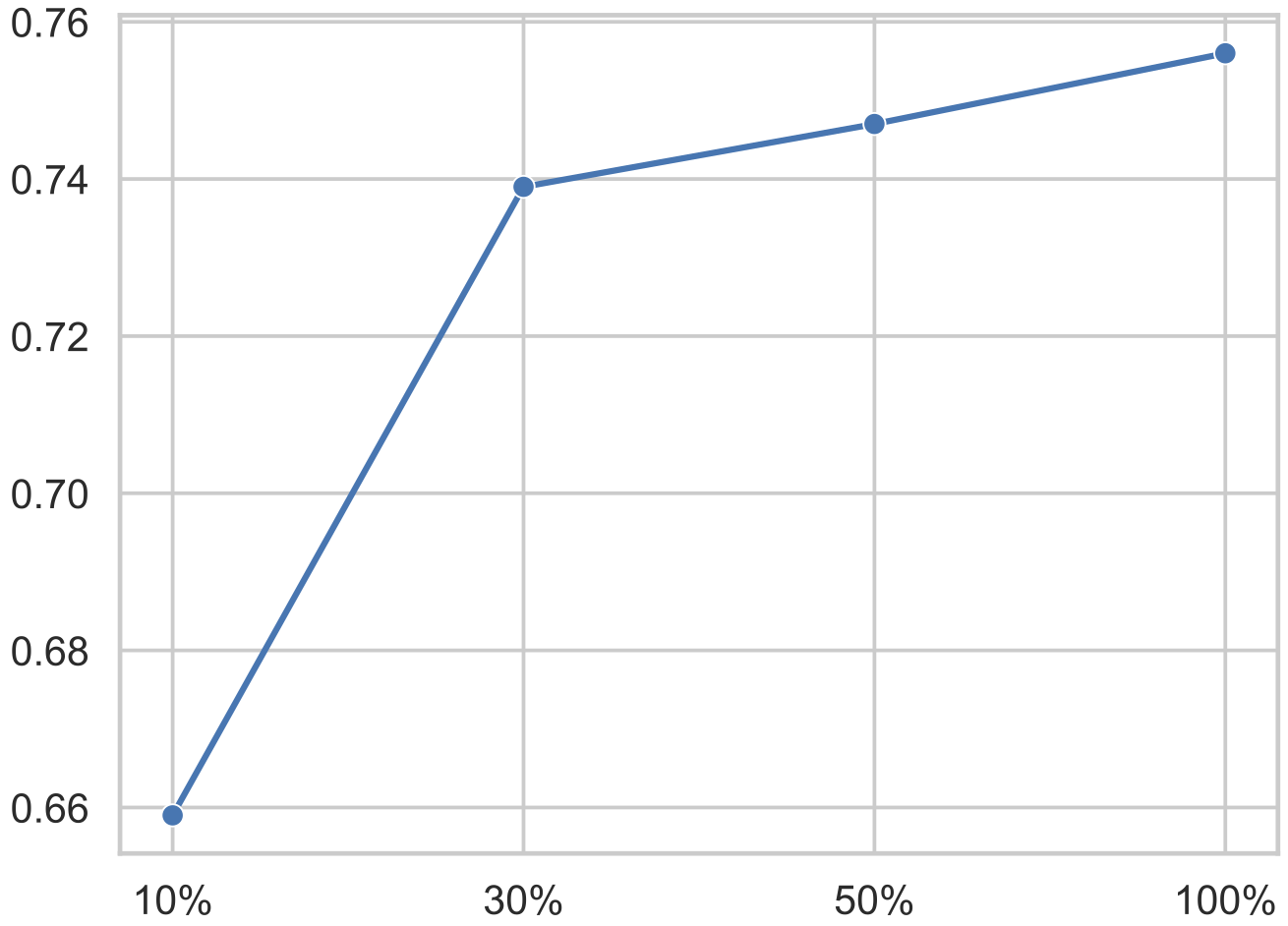}
  \end{center}
  \vspace{-1.em}
  \caption{Averaged pixel-level grounding performance (mIoU) in GemBench as a function of training data size.}
  \label{fig:scaling_curve}
  \vspace{-1.em}
\end{figure}

\noindent\textbf{Training data size.}
Figure~\ref{fig:scaling_curve} shows grounding performance averaged over 4 levels with varying percentages of finetuning data. With just 10\% of the data, the model reaches 87\% of the full-data performance, and performance continues to improve as more data is used.

\begin{table}[t]
\centering
\caption{Offline object grounding comparison on GemBench. `FT' denotes fine-tuning on our synthetic datasets, and `GT obj.' denotes using ground-truth object queries as prior grounding VLMs cannot predict task plans.}
\label{tab:sota_offline_eval}
\begin{tabular}{ccccccc} \toprule
Method & FT & GT obj. & L1 & L2 & L3 & L4 \\ \midrule
RoboPoint~\cite{yuan2024robopoint} & $\times$ & \checkmark & 23.0 & 22.0 & 18.6 & 18.2 \\
LISA~\cite{lai2024lisa} & $\times$ & \checkmark & 25.7 & 24.0 & 19.5 & 24.8 \\ \midrule
LISA~\cite{lai2024lisa} & \checkmark & \checkmark & 62.3 & 56.7 & 37.8 & 43.7 \\
Gondola (Ours) & \checkmark & $\times$ & \textbf{89.5} & \textbf{85.2} & \textbf{73.8} & \textbf{53.8} \\ \bottomrule
\end{tabular}
\vspace{-1.5em}
\end{table}

\subsubsection{Comparison with Grounding VLMs}

Since prior grounding VLMs~\cite{yuan2024robopoint,lai2024lisa} do not directly generate task plans following the prompt, we evaluate their grounding capability independently by querying each model with the ground-truth object name. We select RoboPoint~\cite{yuan2024robopoint} and LISA~\cite{lai2024lisa} as baseline grounding models. RoboPoint takes language instructions as input and predicts sparse keypoints as grounding outputs. To enable quantitative comparison with mask-based methods, we convert the predicted points into segmentation masks using SAM2~\cite{ravi2024sam2}. In contrast, LISA unifies segmentation within the VLM architecture and directly outputs segmentation masks for queried objects.
As existing grounding models do not support multi-view inputs, we apply them independently to each view and compute the mean IoU for evaluation. 

The results are summarized in Table~\ref{tab:sota_offline_eval}. Comparing RoboPoint and LISA, we observe that direct mask prediction via VLM outperforms point-based grounding followed by mask generation. In particular, RoboPoint frequently predicts points outside the target object, which leads to inaccurate masks when processed by SAM2. 
Fine-tuning LISA on our datasets significiantly improves its performance; however, it remains inferior to our approach. 
Notably, our model jointly reasons over multi-view observations and generates pixel-level masks within a unified framework, resulting in substantially stronger grounding performance.
These results validate the effectiveness of our planning-oriented datasets and highlight the advantages of multi-view inputs and pixel-level grounding.

\subsection{Task Completion Evaluation}

\input{tables/gembench_sota_cmpr}

\subsubsection{Comparison with state of the art}
Table~\ref{tab:gembench_sota_cmpr} compares Gondola with state-of-the-art methods on the GemBench test split. 
The methods in the upper section are end-to-end 3D-based policies trained on GemBench L1 tasks without using LLMs.
While these methods perform well on seen tasks (L1), they show limited generalization to novel objects in L2 and L3 and struggle with long-horizon tasks in L4.

The bottom block evaluates models built upon LLM-based backbones including 2D and 3D VLAs.
GR00T~\cite{bjorck2025gr00t} and EO-1~\cite{qu2025eo1} are 2D VLAs fine-tuned on the GemBench training set, using four-view RGB inputs and flow matching for action prediction. 
For EO-1, we freeze the image encoder and fine-tune the remaining components; for GR00T, we fine-tune only the action expert, following the official paper.
Notably, without explicit 3D spatial information, these 2D VLAs face significant performance bottlenecks even at Level 1, despite their extensive pretraining on robot datasets.

BridgeVLA~\cite{li2025bridgevla}, a 3D VLA fine-tuned on GemBench, benefits from large-scale Internet pretraining on image understanding and grounding, and outperforms the other compared methods on L1–L3 tasks. However, it remains limited in long-horizon reasoning.
3D-LOTUS++~\cite{garcia2024gembench} is a modular framework that integrates off-the-shelf LLMs, object detector and segmentation models via prompt engineering and a trained 3D motion-planning policy for execution. While it achieves stronger performance on long-horizon tasks (L4), it performs worse than BridgeVLA on the other levels, because it lacks explicit grounding training.

Compared to prior methods, Gondola's performance is competitive on L1 while establishing new state-of-the-art results across levels L2–L4.
Compared to current end-to-end VLAs, these results validate the effectiveness of a hybrid architecture that couples a grounded VLM for high-level reasoning with a 3D-based policy for precise execution.

\input{tables/gembench_online_ablation}

\subsubsection{Ablations}
We conduct ablation studies to analyze the effect of jointly modeling planning and grounding within a unified architecture, and the impact of closed-loop planning under different action chunk sizes.

\noindent\textbf{Unified planning and grounding.}
Gondola jointly generates textual task plans and pixel-level object grounding within a single model. To evaluate the benefit of this unified formulation, we train two separate models: one dedicated to high-level task planning and another to object grounding. As shown in the first two rows of Table~\ref{tab:gembench_online_eval}, the unified model consistently outperforms the decoupled variants. In addition to improved task success, the unified formulation is more memory- and compute-efficient.

\noindent\textbf{Action chunking and replanning frequency.}
We further analyze the impact of the action chunk size in the closed-loop planning-and-control setup in Table~\ref{tab:gembench_online_eval}. When the chunk size is set to 1, Gondola replans at every execution step; when set to 5, replanning occurs only after the motion policy completes a longer action sequence. 
We observe that tasks requiring fine-grained manipulation benefit from more frequent replanning, as it allows rapid correction of grounding or execution errors. 
In contrast, long-horizon tasks benefit from larger action chunks, which preserve high-level plan consistency. This trade-off highlights the interaction between planning granularity and execution dynamics, and suggests that adaptive replanning strategies may further improve performance.

\input{tables/real_robot}

\subsection{Real Robot Experiments}
\label{sec:suppmat_realrobot}

\noindent\textbf{Experimental setup.}
As illustrated in Figure~\ref{fig:real_robot_setup}, our real robot setup includes three RealSense d435 cameras attached to a table and a 6-DoF UR5 robotic arm equipped with an RG6 gripper. We collect 20 demonstrations per task via teleoperation for 4 tasks. The task instructions are presented in Figure~\ref{tab:real_robot_seen_tasks}.
Then, we fine-tune Gondola on a joint dataset of RLBench and the real robot demonstrations, and combine it with the 3D-based motion planning policy 3D-LOTUS~\cite{garcia2024gembench} in closed-loop control.
For each task, we run models 10 times and report the success rate. 

\noindent\textbf{Real Robot Results.}
Table~\ref{tab:real_robot_seen_tasks} shows the performance on real-world tasks.
We compare against a state-of-the-art VLA model GR00T-N1.5~\cite{bjorck2025gr00t} fine-tuned on our real-robot dataset.
Due to the limited amount of fine-tuning data, GR00T performs noticeably worse than Gondola. GR00T can complete the task when object configurations closely match those seen during training; however, when objects are placed in unseen positions, it frequently misses the target because of inaccurate gripper placement.
Our model, instead, couples high-level grounded task planning with a 3D-based policy, enabling more robust adaptation to novel object arrangements.
Our analysis of Gondola's failure cases reveals that cross-view consistency often degrades in real-world environments, leading to segmentation errors.
We leave this as an important direction for future work, for example, leveraging more real-world multi-view images in training. 

%% file: tables/gembench_offline_eval.tex
\begin{table*}
\centering
\caption{Performance on grounded planning evaluation. We measure the action (Act) and object (Obj) name prediction accuracy and grounding performance (Grd) on the four levels of GemBench validation split. All the models are fine-tuned on the robot grounded planning dataset.}
\label{tab:gembench_offline_eval}
\tabcolsep=0.18cm
\small
\label{tab:gembench_offline_eval}
\begin{tabular}{ccc
>{\columncolor[HTML]{EDEDF5}}c 
>{\columncolor[HTML]{EDEDF5}}c 
>{\columncolor[HTML]{EDEDF5}}c 
>{\columncolor[HTML]{E8F1FA}}c 
>{\columncolor[HTML]{E8F1FA}}c 
>{\columncolor[HTML]{E8F1FA}}c 
>{\columncolor[HTML]{E8E8FA}}c 
>{\columncolor[HTML]{E8E8FA}}c 
>{\columncolor[HTML]{E8E8FA}}c 
>{\columncolor[HTML]{F1E8FA}}c 
>{\columncolor[HTML]{F1E8FA}}c 
>{\columncolor[HTML]{F1E8FA}}c } \toprule
 &  &  & \multicolumn{3}{c}{\cellcolor[HTML]{EDEDF5}L1} & \multicolumn{3}{c}{\cellcolor[HTML]{E8F1FA}L2} & \multicolumn{3}{c}{\cellcolor[HTML]{E8E8FA}L3} & \multicolumn{3}{c}{\cellcolor[HTML]{F1E8FA}L4} \\
\multirow{-2}{*}{\begin{tabular}[c]{@{}c@{}}Grd\\ type\end{tabular}} & \multirow{-2}{*}{\begin{tabular}[c]{@{}c@{}}Multi-\\ view\end{tabular}} & \multirow{-2}{*}{\begin{tabular}[c]{@{}c@{}}Hist-\\ ory\end{tabular}} & Act & Obj & Grd & Act & Obj & Grd & Act & Obj & Grd & Act & Obj & Grd \\ \midrule
Box & \checkmark & \checkmark & 95.1 & 93.7 & 62.8 & 97.4 & 89.0 & 58.7 & 69.3 & 53.8 & 46.2 & 70.0 & 36.8 & 16.6 \\
Mask & $\times$ & $\times$ & 98.0 & 98.2 & 87.8 & 95.1 & 89.5 & 79.8 & 79.3 & 76.3 & 62.7 & 77.2 & 40.0 & 37.4 \\
Mask & \checkmark & $\times$ & \textbf{100} & \textbf{100} & \textbf{88.6} & 98.0 & 91.3 & \textbf{81.2} & 85.6 & 75.6 & 61.4 & \textbf{89.9} & \textbf{50.1} & \textbf{46.5} \\
Mask & \checkmark & \checkmark & \textbf{100} & \textbf{100} & 87.9 & \textbf{99.0} & \textbf{93.3} & 79.2 & \textbf{88.6} & \textbf{83.9} & \textbf{66.4} & 79.4 & 44.9 & 40.0 \\ \bottomrule
\end{tabular}
\end{table*}

%% file: tables/gembench_offline_eval_data.tex
\begin{table*}
\centering
\small
\tabcolsep=0.18cm
\caption{Performance on grounded planning evaluation. All the models use multi-view and history plans for mask generation, but are fine-tuned on different composition of datasets: robot grounded planning (Plan), multi-view referring expression (RefExp), and pseudo long-horizon tasks (Long).}
\label{tab:gembench_offline_eval_data}
\begin{tabular}{ccc
>{\columncolor[HTML]{EDEDF5}}c 
>{\columncolor[HTML]{EDEDF5}}c 
>{\columncolor[HTML]{EDEDF5}}c 
>{\columncolor[HTML]{E8F1FA}}c 
>{\columncolor[HTML]{E8F1FA}}c 
>{\columncolor[HTML]{E8F1FA}}c 
>{\columncolor[HTML]{E8E8FA}}c 
>{\columncolor[HTML]{E8E8FA}}c 
>{\columncolor[HTML]{E8E8FA}}c 
>{\columncolor[HTML]{F1E8FA}}c 
>{\columncolor[HTML]{F1E8FA}}c 
>{\columncolor[HTML]{F1E8FA}}c } \toprule
\multicolumn{3}{c}{Finetuning Data} & \multicolumn{3}{c}{\cellcolor[HTML]{EDEDF5}L1} & \multicolumn{3}{c}{\cellcolor[HTML]{E8F1FA}L2} & \multicolumn{3}{c} {\cellcolor[HTML]{E8E8FA}L3} & \multicolumn{3}{c}{\cellcolor[HTML]{F1E8FA}L4} \\ 
Plan & RefExp & Long & Act & Obj & Grd & Act & Obj & Grd & Act & Obj & Grd & Act & Obj & Grd \\ \midrule
\checkmark & $\times$ & $\times$ & \cellcolor[HTML]{EDEDF5}\textbf{100} & \cellcolor[HTML]{EDEDF5}\textbf{100} & \cellcolor[HTML]{EDEDF5}87.9 & \cellcolor[HTML]{E8F1FA}99.0 & \cellcolor[HTML]{E8F1FA}93.3 & \cellcolor[HTML]{E8F1FA}79.2 & \cellcolor[HTML]{E8E8FA}88.6 & \cellcolor[HTML]{E8E8FA}83.9 & \cellcolor[HTML]{E8E8FA}66.4 & \cellcolor[HTML]{F1E8FA}79.4 & \cellcolor[HTML]{F1E8FA}44.9 & \cellcolor[HTML]{F1E8FA}40.0 \\
\checkmark & \checkmark & $\times$ & \textbf{100} & \textbf{100} & 89.1 & 99.0 & 95.1 & 84.2 & \textbf{92.1} & \textbf{88.2} & 73.3 & 72.1 & 42.3 & 41.7 \\
\checkmark & \checkmark & \checkmark & \multicolumn{1}{r}{\cellcolor[HTML]{EDEDF5}\textbf{100}} & \multicolumn{1}{r}{\cellcolor[HTML]{EDEDF5}\textbf{100}} & \multicolumn{1}{r}{\cellcolor[HTML]{EDEDF5}\textbf{89.5}} & \multicolumn{1}{r}{\cellcolor[HTML]{E8F1FA}\textbf{99.7}} & \multicolumn{1}{r}{\cellcolor[HTML]{E8F1FA}\textbf{95.3}} & \multicolumn{1}{r}{\cellcolor[HTML]{E8F1FA}\textbf{85.2}} & \multicolumn{1}{r}{\cellcolor[HTML]{E8E8FA}88.5} & \multicolumn{1}{r}{\cellcolor[HTML]{E8E8FA}82.2} & \multicolumn{1}{r}{\cellcolor[HTML]{E8E8FA}\textbf{73.8}} & \multicolumn{1}{r}{\cellcolor[HTML]{F1E8FA}\textbf{93.9}} & \multicolumn{1}{r}{\cellcolor[HTML]{F1E8FA}\textbf{51.2}} & \multicolumn{1}{r}{\cellcolor[HTML]{F1E8FA}\textbf{53.8}} \\ \midrule
\end{tabular}
\vspace{-1.5em}
\end{table*}

%% file: tables/gembench_sota_cmpr.tex
\begin{table*}
\centering
\small
\tabcolsep=0.35cm
\caption{Success rate of task execution on four levels of GemBench testing split. }
\label{tab:gembench_sota_cmpr}
\begin{tabular}{llc
>{\columncolor[HTML]{EDEDF5}}c 
>{\columncolor[HTML]{E8F1FA}}c 
>{\columncolor[HTML]{E8E8FA}}c 
>{\columncolor[HTML]{F1E8FA}}c } \toprule
& Method & 3D & L1 & L2 & L3 & L4 \\ \midrule
\multirow{5}{*}{w/o LLM} & Hiveformer~\cite{guhur2023hiveformer} & \checkmark & 60.3$_{\pm 1.5}$ &  26.1$_{\pm 1.4}$ &  35.1$_{\pm 1.7}$ &  0.0$_{\pm 0.0}$ \\
& PolarNet~\cite{chen2023polarnet} &  \checkmark & 77.7$_{\pm 0.9}$ &  37.1$_{\pm 1.4}$ & 38.5$_{\pm 1.7}$ & 0.1$_{\pm 0.2}$\\
& 3D diffuser actor~\cite{ke20243ddifusseractor} & \checkmark & 91.9$_{\pm 0.8}$ &  43.4$_{\pm 2.8}$ & 37.0$_{\pm 2.2}$ & 0.0$_{\pm 0.0}$ \\
& RVT-2~\cite{goyal2024rvt2} &  \checkmark & 89.1$_{\pm 0.8}$ & 51.0$_{\pm 2.3}$ & 36.0$_{\pm 2.2}$ & 0.0$_{\pm 0.0}$ \\ 
& 3D-LOTUS~\cite{garcia2024gembench} & \checkmark & \textbf{94.3$_{\pm 1.4}$} & 49.9$_{\pm 2.2}$ & 38.1$_{\pm 1.1}$ & 0.3$_{\pm 0.3}$ \\ \midrule
\multirow{5}{*}{w/ LLM} 
& GR00T N1.5~\cite{bjorck2025gr00t} & $\times$ & 20.9$_{\pm 1.2}$ & 6.8$_{\pm 1.6}$ & 26.5$_{\pm 1.6}$ & 0.0$_{\pm 0.0}$ \\ 
& EO-1~\cite{qu2025eo1} & $\times$ & 40.5$_{\pm 1.6}$ & 7.7$_{\pm 0.7}$ & 28.5$_{\pm 2.0}$ & 0.0$_{\pm 0.0}$ \\
& BridgeVLA~\cite{li2025bridgevla} & \checkmark & 91.1$_{\pm 1.1}$ & 65.0$_{\pm 1.3}$ & 43.8$_{\pm 1.2}$ & 0.0$_{\pm 0.0}$ \\ 
& 3D-LOTUS++~\cite{garcia2024gembench} & \checkmark & 68.7$_{\pm 0.6}$ & 64.5$_{\pm 0.9}$ & 41.5$_{\pm 1.8}$ & 17.4$_{\pm 0.4}$ \\ 
& Gondola (Ours) & \checkmark & 87.3$_{\pm 1.9}$ & \textbf{74.8}$_{\pm 1.8}$ & \textbf{52.4}$_{\pm 2.1}$ & \textbf{19.0}$_{\pm 1.0}$ \\
\bottomrule
\end{tabular}
\end{table*}

%% file: tables/gembench_online_ablation.tex
\begin{table}
\vspace{-1em}
\centering
\tabcolsep=0.1cm
\small
\caption{Ablation study on unified planning–grounding and action chunk size. Unified modeling improves performance, and replanning frequency affects the trade-off between fine-grained control and long-horizon consistency.}
\label{tab:gembench_online_eval}
\begin{tabular}{cccccc} \toprule
\begin{tabular}[c]{@{}c@{}}Act \\ chunk\end{tabular} & Unified & L1 & L2 & L3 & L4 \\ \midrule
 & $\times$ & 80.8$_{\pm 1.2}$ & 68.5$_{\pm 1.5}$ & 48.6$_{\pm 1.1}$ & 4.1$_{\pm 1.3}$ \\
\multirow{-2}{*}{5} & \checkmark & 87.3$_{\pm 1.9}$ & 74.8$_{\pm 1.8}$ & \textbf{52.4}$_{\pm 2.1}$ & 19.0$_{\pm 1.0}$ \\
\midrule
1  & \checkmark & \textbf{90.8}$_{\pm 1.2}$ & \textbf{78.2}$_{\pm 1.4}$ & 49.5$_{\pm 0.5}$ & 14.9$_{\pm 2.2}$ \\
\bottomrule
\end{tabular}
\vspace{-1.5em}
\end{table}

%% file: tables/real_robot.tex
\begin{figure}[t]
\vspace{-1em}
    \centering
    \begin{subfigure}{0.24\textwidth}
        \centering
        \includegraphics[width=\linewidth]{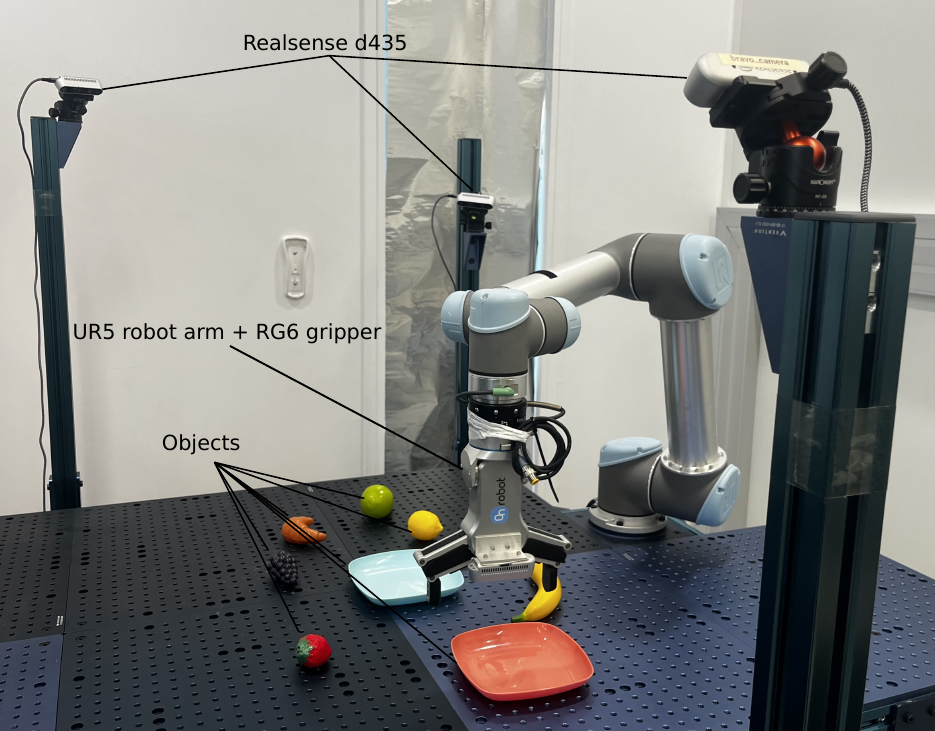}
        \caption{Real robot setup.}
        \label{fig:real_robot_setup}
    \end{subfigure}
    \begin{subfigure}{0.23\textwidth}
        \centering
        \tabcolsep=0.03cm
        \footnotesize

\begin{tabular}{lcc} \toprule
\textbf{} & GR00T & Gondola \\ \midrule
\begin{tabular}[c]{@{}l@{}}Stack yellow cup \\ onto pink cup\end{tabular} & 2/10 & \textbf{8/10} \\
\begin{tabular}[c]{@{}l@{}}Stack navy cup \\ onto yellow cup\end{tabular} & 1/10 & \textbf{7/10} \\
\begin{tabular}[c]{@{}l@{}}Put strawberry \\ in box\end{tabular} & 2/10 & \textbf{4/10} \\
Put peach in box & 2/10 & \textbf{4/10} \\ \midrule
Avg. & 1.8/10 & \textbf{5.8/10} \\  \bottomrule
\end{tabular}

        \caption{Tasks performance.}
        \label{tab:real_robot_seen_tasks}
    \end{subfigure}
   \caption{Real-robot experimental setup and models' performance on real-robot tasks, evaluated by success rate (SR).}
    \label{fig:robot_setup_and_performance}
    \vspace{-1.5em}
\end{figure}

%% file: 10_conclusion.tex
\section{Conclusion}
\label{sec:conclusion}

This paper introduces Gondola, a framework for visually grounded task planning in robotic manipulation. By leveraging multi-view perception and planning history, Gondola generates structured next-step plans enriched with pixel-level segmentation masks. To facilitate effective training,  we curate a diverse dataset encompassing both short- and long-term planning tasks to strengthen temporal reasoning, alongside multi-view referring expressions to enhance object grounding. 
Experimental results on the GemBench benchmark demonstrate that Gondola achieves state-of-the-art performance in both standalone grounded planning and closed-loop task execution. 
Extensive ablations demonstrate the importance of unified multi-view pixel-level grounding and long-horizon planning data for effective high-level reasoning.
Future research will explore scaling these capabilities using large-scale real-world datasets to enhance the model's perceptual robustness in unstructured environments.